\documentclass[lettersize,journal]{IEEEtran}
\usepackage{amsmath,amsfonts}
\usepackage{textcomp}
\usepackage{url}
\usepackage{verbatim}
\usepackage{graphicx}
\usepackage{cite}
\usepackage[utf8]{inputenc} 
\usepackage[T1]{fontenc}    
\usepackage{hyperref}       
\usepackage{url}            
\usepackage{booktabs}       
\usepackage{amsfonts}       
\usepackage{nicefrac}       
\usepackage{microtype}      
\usepackage{xcolor}         
\usepackage{amsmath}
\usepackage{threeparttable}
\usepackage{comment}
\usepackage{enumerate}
\usepackage{varwidth}
\usepackage{array}
\usepackage{bbding}
\usepackage[most]{tcolorbox}
\usepackage{wrapfig}
\usepackage{latexsym}

\newtcolorbox{promptbox}[1][]{
    colback=white,
    colframe=black,
    coltitle=white,
    title=#1,
    sharp corners,
    boxrule=0.5pt,
    left=2mm,
    right=2mm,
    top=1mm,
    bottom=1mm
}

\usepackage{multirow}
\usepackage{graphicx}
\usepackage{caption}
\usepackage{subcaption}
\usepackage{threeparttable}
\usepackage{pifont}
\usepackage{enumitem}
\usepackage{algorithm}
\usepackage{algpseudocode}
\usepackage{multirow}
\usepackage{xcolor}
\usepackage{tcolorbox}
\usepackage{colortbl}
\usepackage{algpseudocode}
\usepackage{arydshln}

\usepackage{booktabs}
\usepackage{tabularx}
\usepackage{array}
\usepackage{makecell}
\usepackage{pifont}

\usepackage{booktabs}
\usepackage{array}
\usepackage{pifont}
\usepackage{xcolor}

\newcolumntype{C}[1]{>{\centering\arraybackslash}m{#1}}

\definecolor{headercolor}{RGB}{220,220,220}
\definecolor{bestcolor}{RGB}{200,255,200}

\newcolumntype{C}[1]{>{\centering\arraybackslash}m{#1}}

\begin{document}

\title{Temporal Evidence-Parameter Co-Alignment for LLM Reasoning over Temporal Knowledge Graphs}

\author{Jiapu Wang
\thanks{Jiapu Wang is with the School of Computer Science and Engineering, Nanjing University of Science and Technology, Nanjing, Jiangsu, 210094, China (e-mail: jiapu.wang@njust.edu.cn).}}

\markboth{Journal of \LaTeX\ Class Files,~Vol.~14, No.~8, August~2021}%
{Shell \MakeLowercase{\textit{et al.}}: A Sample Article Using IEEEtran.cls for IEEE Journals}

\maketitle

\begin{abstract}
Large Language Models (LLMs) often struggle with temporally grounded reasoning because their static parametric memory is misaligned with the dynamic evolution of real-world knowledge. Although Retrieval-Augmented Generation (RAG) and Parameter-Efficient Fine-Tuning (PEFT) provide practical ways to enhance LLMs, they remain insufficient for temporal reasoning. Existing retrieval methods typically prioritize semantic relevance while overlooking temporal validity, event order, and evolutionary dependencies, resulting in evidence that is relevant in isolation but temporally fragmented. Meanwhile, conventional PEFT methods usually employ fixed and input-agnostic low-rank updates, making it difficult to capture time-dependent entity semantics and relation drift. To address these limitations, we propose \underline{T}emporal \underline{E}vidence-\underline{P}arameter \underline{C}o-\underline{A}lignment (TEPCA), a framework for LLM reasoning over Temporal Knowledge Graphs (TKGs). TEPCA aligns LLM reasoning with temporal knowledge at both the evidence and parameter levels. At the evidence level, it introduces time-constrained \textit{Dynamic Relevance-Guided Path (DRGP) pruning} to preserve globally coherent temporal paths, together with \textit{Mutual-Information (MI)-based pruning} to retain query-informative local facts under temporal constraints. At the parameter level, it integrates \textit{Dynamic-Gated LoRA (DG-LoRA)}, a timestamp-conditioned low-rank adaptation mechanism that dynamically modulates the effective adaptation subspace according to temporal context. This design enables LLMs to better capture fine-grained entity evolution and time-dependent relation semantics while preserving parameter efficiency. Experimental results on two TKG benchmarks demonstrate that TEPCA achieves state-of-the-art performance. Additional evaluations of DG-LoRA on eight commonsense QA datasets further verify its general input-adaptive capacity, scalability across LLM backbones, and interpretability.
\end{abstract}


\section{Introduction}

\begin{figure}[h]
\centering
    \includegraphics[scale=0.055]{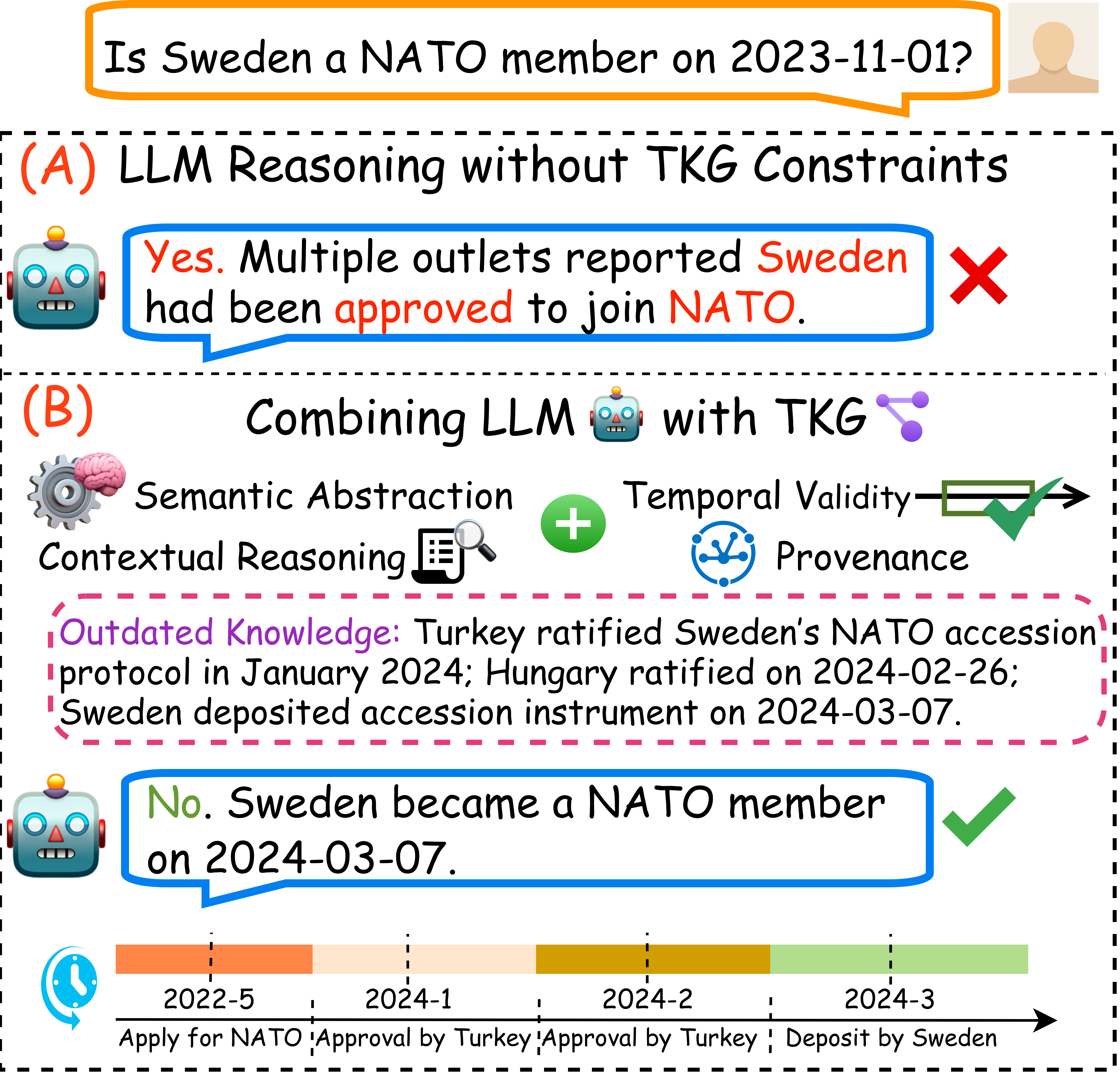}
    \caption{Effect of TKGs constraints on LLM reasoning. }
    \label{fig:my_label_M}
\end{figure}

\IEEEPARstart{L}{arge} Language Models (LLMs) \cite{chang2024survey, zhao2023survey} exhibit emergent capabilities in logical reasoning \cite{chenghaozhu2025your} and generalization \cite{wu2025cdtp}. However, a fundamental architectural paradox persists: while the world is continuously evolving, the parametric memory of LLMs remains frozen at the pre-training stage \cite{wang2024large}. This ``static-dynamic mismatch'' renders state-of-the-art models brittle against temporal drift, often leading to hallucinations where outdated facts are presented as current truths \cite{ouyang2025hoh, bajpai2024temporally}. Since retraining LLMs is computationally prohibitive, researchers often turn to standard Retrieval-Augmented Generation (RAG) \cite{hu2024grag}. Yet, standard RAG retrieves facts based on semantic similarity rather than temporal relevance, resulting in a confusing ``contextual collage'' that fails to capture the evolutionary lineage of events.

Integrating Temporal Knowledge Graphs (TKGs)~\cite{wang2023survey,10.1109/TPAMI.2024.3417451} provides a structured foundation for modeling fact provenance, temporal validity, and event evolution, thereby offering a promising way to mitigate the static--dynamic mismatch of LLMs (Fig.~\ref{fig:my_label_M}). However, the key challenge lies not merely in the availability of external temporal knowledge, but in whether such knowledge can be retrieved, ordered, and utilized under temporal constraints. Existing pipelines, including standard RAG and GraphRAG~\cite{gao2023retrieval,zhang2025survey}, typically prioritize semantic relevance or static graph proximity while overlooking validity intervals, temporal order, and evolutionary dependencies. Consequently, they tend to flatten dynamic histories into static snapshots and retrieve facts that are individually relevant but temporally incompatible with the query, forming a fragmented evidence context rather than a coherent temporal evidence chain~\cite{gade2025s}. Such evidence-level temporal misalignment obscures the evolutionary lineage of events and prevents LLMs from reliably distinguishing outdated states from current truths, allowing hallucinations to persist even when external knowledge is provided.

This limitation becomes more pronounced in time-sensitive reasoning scenarios. Most retrieval-augmented or graph-augmented methods~\cite{peng2024graph,he2024g,hu2024grag} still operate on near-static snapshots and remain insufficiently sensitive to temporal dynamics~\cite{wang2024large}. As illustrated by Lin \textit{et al.}~\cite{lin2025temporal}, snapshot-based methods such as Plan-on-Graph (PoG)~\cite{chen2024plan} can perform reasonably well on slow-changing facts, but degrade markedly on fast-changing ones due to stale parametric knowledge and the lack of explicit temporal constraints~\cite{10.1109/TPAMI.2025.3545671,ime2024}. Similarly, GraphRAG-style pipelines often retrieve short and local chains from near-static KGs, which may be semantically plausible yet chronologically inconsistent~\cite{guo2025empowering}. Without explicitly modeling validity windows, temporal precedence, and cross-evidence consistency, the retrieved paths may conflict with the query timestamp or with one another, thereby weakening factual grounding and exacerbating hallucinations caused by temporal inconsistency~\cite{dhingra2022time,huang2025survey}.

Nevertheless, evidence-level alignment alone is insufficient for temporally grounded reasoning, because the LLM must also internalize and utilize time-dependent semantics through its adaptation mechanism. Mainstream Parameter-Efficient Fine-Tuning (PEFT) methods, such as LoRA~\cite{hu2022lora} and DoRA~\cite{liu2024dora}, are computationally efficient and effective in many downstream tasks, but they usually employ fixed and input-agnostic low-rank updates. This design implicitly assumes that entity and relation semantics remain stable across contexts, which is often violated in evolving knowledge environments. For instance, the geopolitical role and relational context of \texttt{Germany} may vary substantially across historical periods, and such period-specific semantics are difficult to capture with a shared time-invariant adaptation subspace~\cite{taheri2019learning}. By forcing heterogeneous temporal states into the same static low-rank space, conventional PEFT may induce a form of ``temporal specificity collapse'', where time-specific signals are over-smoothed or incorrectly transferred across timestamps. Consequently, the model may propagate outdated temporal semantics and generate predictions that are inconsistent with the query-specific temporal context.

To address these challenges, we propose TEPCA (\textbf{T}emporal \textbf{E}vidence-\textbf{P}arameter \textbf{C}o-\textbf{A}lignment), a framework that aligns LLM reasoning with temporal knowledge at both the evidence and parameter levels. At the evidence level, TEPCA retrieves temporally grounded subgraphs from TKGs by jointly considering semantic relevance, temporal validity, and evolutionary structure. Specifically, it introduces \textit{Dynamic Relevance-Guided Path (DRGP) pruning} to preserve globally coherent temporal paths that reflect long-range event evolution, and \textit{Mutual Information--based (MI-based) pruning} to retain query-informative local facts under temporal constraints. These two complementary pruning strategies allow the model to access evidence that is not only relevant to the query but also consistent with its temporal context.

At the parameter level, TEPCA introduces Dynamic-Gated LoRA (DG-LoRA), a timestamp-conditioned adaptation mechanism designed to overcome the static nature of conventional low-rank fine-tuning. Instead of applying the same low-rank update across all temporal contexts, DG-LoRA dynamically modulates the effective adaptation subspace according to temporal signals such as query time and validity windows. This enables the model to allocate adaptation capacity in an input- and time-dependent manner, thereby capturing fine-grained semantic drift in evolving entities and relations. By jointly aligning what evidence should be retrieved, when the evidence is valid, and how the LLM should adapt to time-dependent semantics, TEPCA provides a principled framework for accurate, interpretable, and temporally grounded LLM reasoning over dynamic knowledge.

The main contributions of this paper are summarized as follows:

\begin{itemize}[left=0cm]
    \item This paper proposes a \textbf{T}emporal \textbf{E}vidence-\textbf{P}arameter \textbf{C}o-\textbf{A}lignment (TEPCA) framework, which effectively leverages temporal dynamics within Temporal Knowledge Graphs (TKGs) for enhancing Large Language Models (LLMs) reasoning, while significantly improving interpretability;
    \item This paper proposes a novel temporal graph retrieval strategy that combines Dynamic Relevance-Guided Path (DRGP) pruning for global subgraph optimization with Mutual Information (MI)-based filtering to extract query-relevant details;
    \item This paper designs Dynamic-Gated LoRA, a time-aware adaptation mechanism that dynamically adjusts the low-rank capacity of LLMs based on temporal context, enabling fine-grained reasoning over temporally evolving knowledge.
\end{itemize}

\begin{figure*}[t]
\centering
    \includegraphics[scale=0.095]{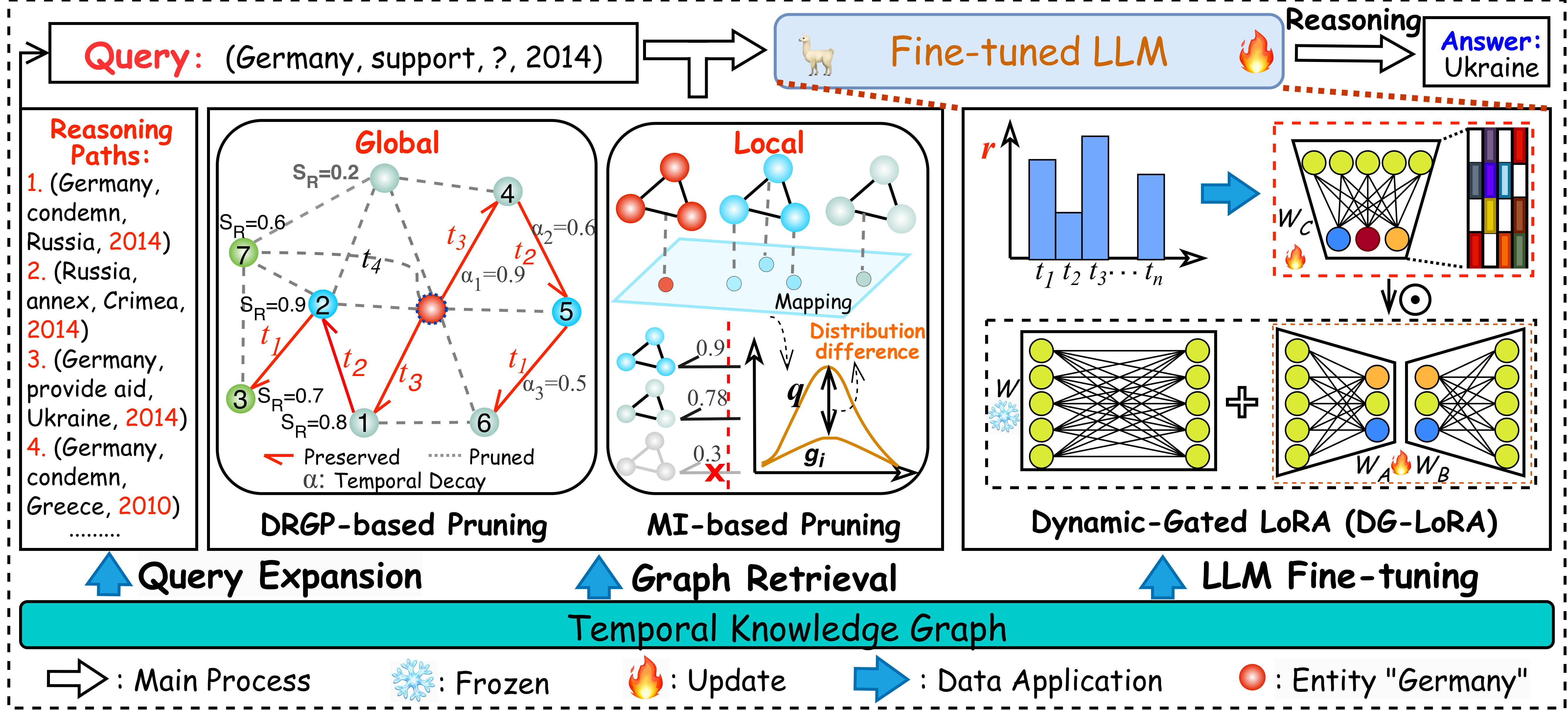}
    \caption{The framework of TEPCA. TEPCA fine-tunes LLMs via Dynamic-Gated LoRA (DG-LoRA), which injects time-adaptive low-rank updates modulated by query-conditioned gating. Query Expansion then identifies temporally relevant event chains to enhance contextual understanding. The Graph Retrieval module applies Dynamic Relevance-Guided Path (DRGP) Pruning and Mutual Information (MI)–based Pruning to distill globally consistent paths and locally informative facts from TKGs. Global paths are selected based on reverse chronological constraints (e.g., $t_i < t_3$ and $t_3 > t_2 > t_1$), semantic relevance, and temporal decay, ensuring temporal consistency and causal validity. Finally, LLM Reasoning integrates the retrieved temporal subgraph into the fine-tuned LLM to produce time-sensitive and context-aware predictions.}
    \label{fig:my_label_1}
\end{figure*}

\section{Related Work}

\subsection{Temporal Knowledge Graph Reasoning}
Temporal Knowledge Graph Reasoning (TKGR) aims to reason over temporal data within Temporal Knowledge Graphs (TKGs) to predict unobserved relations or infer future events. Existing TKGR methods can be broadly categorized into rule-based and embedding–based approaches. 

\textbf{Rule-based TKGR methods} utilize logical rules and temporal constraints to infer dynamic relations between entities, thereby providing highly interpretable reasoning processes \cite{zhuo2025effective,ijcai2025p413}. TLogic \cite{liu2022tlogic} extracts logical rules from TKGs using temporal random walk methods for predicting future events. Building on this, TILP \cite{xiong2023tilp} proposes a constrained random walk technique and models different types of temporal features to optimize the learning process of temporal logical rules. Further enhancing rule discovery, TR-Rules \cite{li2023tr} employs a lightweight algorithm to reduce temporal redundancy and mine acyclic rules from TKGs. 

Other approaches focus on refining logical reasoning or integrating broader patterns. For instance, LGevo \cite{luo2025temporal} jointly models local and global evolutionary patterns via dynamic subgraph sampling and a temporal fusion encoder, which can inform the formulation of temporal rules. Meanwhile, TFLEX \cite{lin2022tflex} achieves efficient reasoning on TKGs by combining fuzzy logic with temporal logic operators. Conversely, TPRG \cite{bai2023multi} infers new facts by analyzing the logical relationships of multi-hop paths in TKGs and applying fourteen manually defined temporal logical rules. In a different direction, LogiQ \cite{chen2025enhancing} encodes temporal logic rules and applies rule-guided reward shaping to steer reinforcement learning agents toward rule-consistent actions, bridging logical reasoning with reinforcement learning. TLmod \cite{bai2023temporal} introduces a pruning strategy that calculates confidence scores for rules, selecting only those with high confidence for the reasoning on TKGs. While rule-based TKGR methods are highly interpretable, they may lack flexibility when dealing with complex or sparse data.

\textbf{Embedding-based TKGR methods} \cite{DBLP:conf/kdd/DengRN20, wang2023qdn} utilize embedding techniques \cite{tdn2023,multi2023,multilevel2023} to learn high-dimensional representations of entities and relations, capturing the semantic and temporal features of TKGs, thereby enabling the prediction of future events and the inference of unknown relations. RE-NET \cite{jin2019recurrent} employs Recurrent Neural Networks (RNNs) and Graph Neural Networks (GNNs) to model historical facts, thereby enhancing the prediction of future events in TKGs. Building on this, RE-GCN \cite{li2021temporal} predicts future events by leveraging a combination of Relational Graph Convolutional Networks (R-GCN) and Gated Recurrent Units (GRU) to encode historical facts. TiRGN \cite{li2022tirgn} simultaneously integrates local and global encoders to comprehensively explore historical development patterns, while HGLS \cite{DBLP:conf/www/ZhangX0WW23} captures semantic and temporal dependencies at both sub-graph and global graph levels \cite{wang2026comgrl}. Addressing the interpretability challenge, xERTE \cite{han2020explainable} emerges as a pioneering explainable TKGR method, integrating causal-aware relational attention, interpretable inference graphs, and dynamic pruning techniques. CENET \cite{cenet} further enhances reasoning capabilities by integrating both historical and non-historical event data to model robust entity distributions and employing a contrastive learning algorithm to identify key entities. While deep learning-based TKGRs are typically better suited for handling large-scale and complex data, they often lack straightforward interpretability.

\subsection{Large Language Model Reasoning}
Large Language Models (LLMs) reasoning on TKGs \cite{chu2023timebench, wang2023survey} typically involves leveraging TKGs to enhance the generative capabilities of LLMs for reasoning tasks. Existing LLM reasoning methods can be broadly classified into two categories: prompt learning-based and fine-tuning-based reasoning.

\textbf{Prompt Learning-based LLM Reasoning} focuses on improving reasoning abilities using contextual information, customized prompts, or zero-shot learning, without requiring fine-tuning of the LLMs. PPT~\cite{xu2023pre} introduces specialized prompts tailored for different temporal interval types, enabling effective extraction of timestamp semantics. 
HFL~\cite{xu2025historical} adopts a token-only input paradigm to model time intervals, integrating both short-term and long-term historical contexts for robust temporal reasoning. In-context learning (ICL) strategies are explored by GPT-NeoX~\cite{lee2023temporal}, which enhances temporal predictions without fine-tuning by leveraging structured examples. 
Mixtral-8x7B-CoH~\cite{xia2024enhancing} applies a Chain-of-History (CoH) mechanism to model higher-order historical dependencies in a progressive manner. 
Similarly, zrLLM~\cite{ding2024zrllm} investigates zero-shot relational reasoning with LLMs for temporal fact completion.

To further improve temporal constraint modeling, TimeR$^4$~\cite{qian2024timer4} introduces a two-stage retrieve-then-reason framework. 
It first extracts temporal constraints via a retrieve-rewrite module and then identifies relevant evidence using a retrieve-rerank module. 
Most recently, LLM-DA~\cite{wang2024large} proposes a dynamic adaptation strategy that incrementally updates the LLM based on recent events, enhancing generalization for future event prediction.

\textbf{Fine-tuning-based LLM Reasoning} enhances the performance on temporal reasoning tasks by fine-tuning LLMs. Luo \textit{et al.} \cite{luo2024chain} utilizes historical chains to fine-tune LLMs, improving their ability to predict future events. GenTKG \cite{liao2024gentkg} incorporates few-shot parameter-efficient instruction-tuning and a temporal logical rule-based retrieval strategy to enhance the reasoning ability and interpretability of LLMs. ChapTER \cite{peng-etal-2024-deja} is a prefix-tuning-based contrastive historical modeling approach for TKGR that balances textual and temporal information through virtual time prefixes and pseudo-Siamese encoders. TKGIG \cite{yuan2024back} introduces an explainable temporal reasoning task to predict future events, using the ExplainTemp dataset generated via a KG-guided strategy. 

Nonetheless, existing approaches remain largely opaque, offering limited interpretability and making the reasoning process difficult to trace. Additionally, most adopt uniform low-rank adaptation schemes that fail to account for the temporal dynamics of evolving entities. These shortcomings substantially undermine the robustness and real-world applicability of LLMs in producing temporally grounded and explainable outputs.

\section{Preliminary}
\paragraph{Temporal Knowledge Graph (TKG)} A TKG is defined as a tuple $\mathcal{G} = \{ \mathcal{E}, \mathcal{R}, \mathcal{T}, \mathcal{Q} \}$, where $\mathcal{E}$ represents the set of entities, $\mathcal{R}$ denotes the set of relations, $\mathcal{T}$ is the set of timestamps, and $\mathcal{Q}$ comprises the set of quadruplets. Each quadruplet $(e_s, r, e_o, t) \in \mathcal{Q}$ consists of a subject entity $e_s$, an object entity $e_o$, a relation $r$, and a timestamp $t$.

\paragraph{Graph Retrieval-Augmented Generation (GraphRAG)} GraphRAG \cite{peng2024graph} improves the reasoning capability of LLMs by incorporating a graph retriever $\mathcal{M}_{\theta1}(G^{*} \mid q, \mathcal{G})$, which retrieves temporally and contextually relevant KGs in response to a query $q$. The retrieved TKGs are then processed by LLMs $\mathcal{M}_{\theta2}(A \mid q, G^{*})$ to generate the answer:
\begin{equation} 
\mathcal{M}(A \mid q, \mathcal{G}) = \sum_{G^{*} \subseteq \mathcal{G}} \mathcal{M}_{\theta2}(A \mid q, G^{*}) \cdot \mathcal{M}_{\theta1}(G^{*} \mid q, \mathcal{G}),
\end{equation}
where $A$ is the generated answer, $\mathcal{G}$ and $G^{*}$ denote the original and retrieved TKGs, respectively. 

\paragraph{Low-Rank Adaptation (LoRA)} LoRA~\cite{hu2022lora} is a parameter-efficient fine-tuning method designed to adapt LLMs without updating their original weights. Rather than updating the full-rank weight matrices, LoRA introduces trainable low-rank components into selected layers, substantially reducing the number of trainable parameters while maintaining task performance. LoRA can be formulated as:
\begin{equation}
  \begin{aligned}
    y = xW + xW_{A}W_{B},
  \end{aligned}
  \label{eq:LoRA}
\end{equation}
where $x\in \mathbb{R}^{S \times d_1}$ and $y\in \mathbb{R}^{S \times d_2}$ are the input and output of the LLM, $S$ denotes the token sequence length, and $d_1$, $d_2$ are the input and output channel dimensions. The pre-trained weight $W\in \mathbb{R}^{d_1 \times d_2}$ is fixed (non-trainable), $W_{A}\in \mathbb{R}^{d_1 \times r}$ is initialized via uniform Kaiming distribution~\cite{he2015delving}, while $W_{B}\in \mathbb{R}^{r \times d_2}$ is zero-initialized. The rank $r$ satisfies $r \ll \min(d_1, d_2)$ and is shared across samples.

\section{Dynamic-Gated LoRA Fine-tuning}\label{dglora}
TKGs demand models that adapt to entities whose roles and relations evolve over time. Conventional LoRA, which applies the same low‑rank updates uniformly, presumes static semantics and cannot capture this temporal–semantic variability. To overcome these limitations, we propose DG‑LoRA, a novel fine‑tuning framework that conditions low‑rank adaptations on TKGs. DG‑LoRA incorporates a lightweight gating network that dynamically adjusts both the magnitude and the rank of LoRA updates according to the timestamp of each event. This mechanism empowers the model to specialize its internal representations for different time steps, markedly enhancing its ability to reason over evolving knowledge.

DG‑LoRA enhances standard LoRA by incorporating a time-aware gating mechanism into the low-rank adaptation. Specifically, we extend the input to include timestamps: for each token representation \(x_{\text{tok}}\) associated with a timestamp \(t\), the combined input is \(x = [x_{\text{tok}}; t]\). The formulation of DG-LoRA is given as follows:
\begin{equation}
    y \;=\; xW \;+\; \bigl(xW_A \,\odot\, \underbrace{\bigl[(xW_C \ge 0)\,\odot\,\tanh(xW_C)\bigr]}_{\text{gate}_c}\bigr)\,W_B,
    \label{eq:DG-loRA}
\end{equation}
where \(W\) is the frozen pretrained weight matrix, and \(W_A \in \mathbb{R}^{d_1 \times r}\) and \(W_B \in \mathbb{R}^{r \times d_2}\) parameterize the static low‐rank update. The learnable matrix \(W_C\in\mathbb{R}^{d_1\times r}\) generates the gate features:
\begin{equation}
    \text{gate}_c = (xW_C \ge 0)\,\odot\,\tanh(xW_C),
    \label{eq:DG-loRA1}
\end{equation}
where the hard threshold \((xW_C \ge 0)\) induces token-wise sparsity and \(\tanh(xW_C)\) provides continuous scaling. By conditioning the gate on both the token and timestamp, DG-LoRA dynamically adjusts both the magnitude and the effective rank of the adaptation, allowing the model to adapt its low-rank subspace to different temporal contexts. This equips the model with temporally grounded reasoning capabilities necessary for handling evolving knowledge in TKGs. A detailed gradient analysis is presented in Subsection~\ref{DG-LoRA}, with the theoretical justification for DG-LoRA provided in Subsection~\ref{proof}.

\section{Methodology}\label{repre}
In this section, we propose a \textbf{K}nowledge \textbf{G}raph–Grounded \textbf{R}etrieval and \textbf{T}emporal \textbf{A}daptation (TEPCA) framework. As
displayed in Fig. \ref{fig:my_label_1}, the core of our proposed method TEPCA mainly contains three components:
\begin{itemize}[left=0pt]
\item \textit{Query Expansion} incorporates relevant historical event chains to enrich the temporal context of the query;
\item \textit{Graph Retrieval} leverages Dynamic Relevance-Guided Path Pruning and Mutual Information-based pruning algorithms to effectively extract the most relevant global and local information; \item \textit{LLM Reasoning} integrates the retrieved temporal knowledge into the LLM to generate context-aware and time-sensitive responses. 
\end{itemize}

\subsection{Query Expansion}
Temporal queries are often underspecified, lacking the historical context necessary for reliable inference. This contextual sparsity can result in ambiguous or incomplete reasoning, especially in dynamic environments where entities and relations evolve over time. To address this issue, we introduce a query expansion mechanism that enriches the input query with temporally grounded events from the past.
This augmentation enables LLMs to access temporally relevant context, thereby facilitating more accurate and context-aware reasoning. 

Given the edge $r(e_x, e_y, t_l)$, query expansion explores the temporal random walks \cite{nguyen2018dynamic} to retrieve multi-hop relational paths that involve the entity $e_x$ and occurred prior to timestamp $t_l$. To ensure both temporal relevance and causal consistency, TEPCA constrains the random walk process by integrating both temporal order and interval-based decay into a unified transition probability $P_{xy}(t)$:
\begin{equation}
P_{xy}(t) = \frac{\mathbb{I}[t < t_l] \cdot \exp(-\lambda(t_l - t))}{\sum\nolimits_{(e_x,e_y',t') \in \mathcal{G}} \mathbb{I}[t' < t_l] \cdot \exp(-\lambda(t_l - t'))},
\label{eq:temporal_transition}
\end{equation}
where $\mathbb{I}[\cdot]$ ensures only past events are considered, and the exponential decay term $\exp(-\lambda(t_l - t))$ favors more recent interactions relative to the query timestamp. The decay coefficient $\lambda$ controls how strongly temporal distance influences the transition likelihood. 

Through query expansion, we obtain a set of historical events that capture the temporal context and semantics relevant to the query. Each retrieved historical fact in $\mathcal{P}_{\text{exp}}$ is first linearized into a natural language sentence and subsequently encoded by a Pretrained Language Model (PLM). The enhanced query representation $\mathbf{s}_q$ is then derived via mean pooling:
\begin{equation}
\mathbf{s}_q = \frac{1}{|\mathcal{P}_{\text{exp}}|} \sum_{i=1}^{|\mathcal{P}_{\text{exp}}|} \text{PLM}(\mathcal{P}_{\text{exp}}[i]),
\end{equation}
where $\text{PLM}(\mathcal{P}_{\text{exp}}[i])$ denotes the contextual embedding of the $i$-th linearized event. This aggregation mechanism synthesizes the semantic information from historical evidence, yielding a query representation that is both temporally grounded and contextually enriched.


\subsection{Graph Retrieval}
Graph retrieval extracts both global and local temporal contexts from TKGs to support accurate reasoning. Due to the evolving nature of entities and relations in TKGs, effective subgraph extraction must account for both structural relevance and temporal specificity. To this end, the graph retrieval pipeline consists of two complementary pruning strategies: \textit{Dynamic Relevance-Guided Path Pruning} efficiently identifies multi-hop paths that encode global temporal-relational patterns; and \textit{Mutual Information–Based Pruning} leverages mutual information to select the subset of quadruplets that best align semantically and temporally with the query. 

\paragraph{Dynamic Relevance-Guided Path Pruning}
Dynamic Relevance-Guided Path (DRGP) pruning aims to extract globally coherent temporal paths from the query-induced candidate graph $\mathcal{G}_q$. 
Given a temporal query $q=(e_s,r,?,t_q)$, we first restrict the candidate space to historical facts:
\begin{equation}
\mathcal{G}_q^{-}
=
\{g=(e_u,r_g,e_v,t_g)\in\mathcal{G}_q \mid t_g<t_q\}.
\label{eq:drgp_candidate}
\end{equation}

This query-time constraint prevents post-query evidence from being used during path construction. 
Starting from the head entity $e_s$, DRGP expands candidate paths within $\mathcal{G}_q^{-}$ by jointly considering query relevance, path-context compatibility, and temporal order.

Unlike fact-level pruning, DRGP treats a reasoning path as an ordered temporal structure. 
For a candidate path $R=[g_1,g_2,\ldots,g_i]$, where $g_j=(e_j,r_j,e_{j+1},t_j)$, we enforce the reverse chronological constraint:
\begin{equation}
t_q > t_1 > t_2 > \cdots > t_i,
\label{eq:chronological_constraint}
\end{equation}
where this constraint encourages the retrieved path to trace supporting evidence backward from the query timestamp to earlier events.

At the $j$-th expansion step, let $R_{<j}=[g_1,\ldots,g_{j-1}]$ denote the existing partial path. 
We encode the path context as:
\begin{equation}
\mathbf{h}_{R_{<j}}
=
\operatorname{Pool}
\left(
\mathbf{e}_{g_1},
\mathbf{e}_{g_2},
\ldots,
\mathbf{e}_{g_{j-1}}
\right),
\label{eq:path_context}
\end{equation}
where $\mathbf{e}_{g_l}$ is the representation of the linearized quadruplet $g_l$. 
For the first hop, we set $\mathbf{h}_{R_{<1}}=\mathbf{s}_q$, where $\mathbf{s}_q$ denotes the expanded query representation.

To guide path expansion, we define a transition-aware relevance score:
\begin{equation}
\begin{split}
\Psi&(g_j \mid q,R_{<j})
=\ \beta_q\, \cos(\mathbf{s}_q,\mathbf{e}_{g_j}) \\
&+ \beta_p\, \cos(\mathbf{h}_{R_{<j}},\mathbf{e}_{g_j}) + \beta_r\, \cos(\mathbf{r},\mathbf{r}_{g_j}),
\end{split}
\label{eq:drgp_transition}
\end{equation}
where $\mathbf{r}$ and $\mathbf{r}_{g_j}$ denote the representations of the query relation and the candidate relation, respectively. 
The coefficients $\beta_q$, $\beta_p$, and $\beta_r$ balance query relevance, path-context coherence, and relation-level compatibility. 
Different from ordinary semantic retrieval, this score evaluates whether a candidate fact is suitable as the next hop of the current reasoning path, rather than only measuring its isolated relevance to the query.

To model temporal proximity, we assign each candidate fact a decay weight:
\begin{equation}
\alpha_j
=
\exp\left(-\lambda_1(t_q-t_j)\right),
\label{eq:drgp_temporal_decay}
\end{equation}
where $\lambda_1>0$ controls the preference for recent historical facts. 
The score of a partial path $R=[g_1,g_2,\ldots,g_i]$ is then defined as:
\begin{equation}
S_{\mathrm{DRGP}}(R)
=
\frac{
\sum_{j=1}^{i}
\alpha_j
\Psi(g_j \mid q,R_{<j})
}{
\sum_{j=1}^{i}
\alpha_j
},
\label{eq:drgp_path_score}
\end{equation}
where the normalization mitigates length bias and allows paths with different lengths to be compared fairly.

After at most $L$ expansion steps, DRGP selects the top-$k$ paths satisfying Eq.~\eqref{eq:chronological_constraint}:
\begin{equation}
\mathcal{R}_{\mathrm{Top}\text{-}k}
=
\operatorname{TopK}_{R\in\mathcal{R}_{\mathrm{all}}}
S_{\mathrm{DRGP}}(R),
\label{eq:drgp_topk}
\end{equation}
where $\mathcal{R}_{\mathrm{all}}$ denotes all candidate paths constructed from $\mathcal{G}_q^{-}$. 
The optimized global subgraph is obtained by aggregating the selected paths:
\begin{equation}
\mathcal{G}_q^{\mathrm{opt}}
=
\bigcup_{R\in\mathcal{R}_{\mathrm{Top}\text{-}k}} R .
\label{eq:drgp_subgraph}
\end{equation}

DRGP therefore performs path-level pruning to preserve query-relevant, path-coherent, and chronologically valid global evidence. 
This global evidence complements the subsequent MI-based pruning, which performs fact-level selection to retain locally informative evidence from the same temporal candidate space.

\paragraph{Mutual Information-based Pruning}
While DRGP pruning preserves globally coherent temporal paths, path-level selection may overlook local facts that are individually critical to the query. 
To capture such fine-grained evidence, we further prune the query-induced subgraph $\mathcal{G}_q$ by selecting quadruplets with high query-specific information gain. 
Since the joint distribution between natural-language queries and temporal quadruplets is unavailable and difficult to estimate from sparse TKG observations, directly computing the mutual information $I(q;g)$ is intractable. 
We therefore approximate it using a temporal-aware contrastive pointwise mutual-information estimator defined over the finite candidate set $\mathcal{G}_q$.

Given a query $q=(e_s,r,?,t_q)$ and a candidate quadruplet $g=(e_s',r',e_o',t_g)\in\mathcal{G}_q$, we utilize a temperature-scaled cosine similarity to compute the semantic compatibility between $q$ and $g$:
\begin{equation}
\phi(q,g)=\frac{\cos(\mathbf{s}_q,\mathbf{e}_g)}{\tau},
\end{equation}
where $\tau>0$ controls the sharpness of the compatibility distribution.

To incorporate temporal proximity, we introduce a time-decay weight:
\begin{equation}
    \omega(q,g)=\exp\left(-\lambda_2(t_q-t_g)\right),
\end{equation}
where $\lambda_2$ controls the strength of temporal decay. 
This design favors more recent historical facts, while earlier facts can still be retained if they exhibit strong semantic compatibility.

Combining semantic compatibility with temporal proximity, we define a query-conditioned evidence distribution over $\mathcal{G}_q^{-}$:
\begin{equation}
    p_{\theta}(g\mid q)=
    \frac{\omega(q,g)\exp(\phi(q,g))}
    {\sum_{g'\in\mathcal{G}_q^{-}}\omega(q,g')\exp(\phi(q,g'))},
\end{equation}
where it assigns larger probabilities to facts that are both semantically compatible with the query and temporally close to the query timestamp.

However, structurally central facts or high-frequency relations can be over-selected even when they carry limited query-specific information. 
To correct this background bias, we introduce a query-independent background distribution:
\begin{equation}
    p_{0}(g)=
    \frac{\rho(g)}
    {\sum_{g'\in\mathcal{G}_q^{-}}\rho(g')},
\end{equation}
where $\rho(g)$ denotes the background salience of $g$, i.e., its likelihood of being selected without conditioning on query semantics. 
For $g=(e_s',r',e_o',t_g)$, we instantiate $\rho(g)$ as:
\begin{equation}
\rho(g)=\widetilde{\mathrm{deg}}(e_s')+\widetilde{\mathrm{deg}}(e_o')+\eta\,\widetilde{\mathrm{freq}}(r')+\epsilon,
\end{equation}
where $\widetilde{\mathrm{deg}}(\cdot)$ denotes the normalized entity degree in the candidate temporal subgraph, $\widetilde{\mathrm{freq}}(r')$ is the normalized frequency of relation $r'$, $\eta$ balances entity-level and relation-level salience, and $\epsilon$ is a smoothing constant for numerical stability. 

The local information score is then defined as the log-ratio between the query-conditioned evidence distribution and the background distribution:
\begin{equation}
    S_{\mathrm{MI}}(q,g)
    =
    \log
    \frac{p_{\theta}(g\mid q)}{p_{0}(g)}.
\end{equation}
where $S_{\mathrm{MI}}(q,g)$ provides a temporal-aware contrastive approximation of pointwise mutual information, quantifying the information gain of a candidate fact after conditioning on the query and its timestamp relative to its background salience. 
For a query $q$, the ranking induced by $S_{\mathrm{MI}}(q,g)$ is equivalent to:
\begin{equation}
    S_{\mathrm{MI}}(q,g)
    \propto
    \phi(q,g)
    +\log\omega(q,g)
    -\log\rho(g).
\end{equation}
Thus, the proposed score selects facts that are semantically aligned with the query, temporally close to the query timestamp, and informative beyond what is expected from their background salience.

Finally, we construct the local evidence set by selecting the top-$k$ quadruplets with the highest information scores:
\begin{equation}
    \mathcal{V}_q^{\mathrm{local}}
    =
    \mathrm{Top}_k
    \left(
    \left\{
    S_{\mathrm{MI}}(q,g)
    \mid g\in\mathcal{G}_q
    \right\}
    \right).
\end{equation}
Through the dual-pruning strategy, MI-based pruning complements DRGP pruning: DRGP captures long-range temporal structures through globally coherent paths, whereas MI-based pruning retains fine-grained local evidence that directly supports precise answer localization.



\subsection{LLM Reasoning}
LLM Reasoning empowers LLMs to perform temporally grounded and semantically coherent reasoning by integrating localized evidence and globally structured context extracted from TKGs. It is composed of two key components:  
\textit{Graph Encoding}, which encodes the optimized global subgraph into dense representations that preserve both temporal order and relational dependencies; and \textit{Answer Generation}, where the LLM leverages both the global subgraph embedding and the pruned local facts to infer accurate answers aligned with the query's intent.

\paragraph{Graph Encoding}
To incorporate global contextual knowledge into LLMs, we first encode the optimized subgraph $\mathcal{G}_q^{\text{opt}}$ using a graph encoder \cite{velivckovic2017graph}. The encoder captures both relational and temporal dependencies among entities, generating node-level embeddings that retain the subgraph’s structural and temporal characteristics. The resulting node embeddings are aggregated using mean pooling (M-POOL) to form a holistic representation of the subgraph. This representation is then projected into the LLM’s input space through a multilayer perceptron (MLP), ensuring compatibility with the LLM’s embedding space. Formally, this encoding process is defined as:
\begin{equation}
\mathbf{h}_q^{\text{global}} = \text{MLP}_{\phi_2}\big(\text{M-POOL}\big(\text{GNN}_{\phi_1}(\mathcal{G}_q^{\text{opt}})\big)\big),
\end{equation}
where $\text{GNN}_{\phi_1}(\cdot)$ denotes the graph encoder with parameters $\phi_1$, and $\text{MLP}_{\phi_2}(\cdot)$ maps the pooled vector to the LLM-compatible space. The resulting embedding $\mathbf{h}_q^{\text{global}}$ captures rich temporal and relational semantics from the global subgraph.

\paragraph{Answer Generation}
In the final step, the LLM generates an answer $\mathcal{A}_q$ by jointly conditioning on the original query $q$, the encoded global representation $\mathbf{h}_q^{\text{global}}$, and the set of local quadruplets $\mathcal{V}_q^{\text{local}}$. 

These components are embedded and formatted as part of the input prompt, allowing the LLM to perform context-aware reasoning. 
The whole process is defined as:
\begin{equation}
\begin{split}
    \mathcal{M}_{\theta}(\mathcal{A}_q \mid q, \mathcal{G}) &= \sum_{G^{*} \in \mathcal{S}_q} \underbrace{\mathcal{M}_{\theta3}(\mathcal{A}_q \mid q, \mathbf{h}_q^{\text{global}}, \mathcal{V}_q^{\text{local}})}_{\text{LLM Reasoning}}\\ &\cdot \underbrace{\mathcal{M}_{\theta2}(\mathbf{h}_q^{\text{global}}, \mathcal{V}_q^{\text{local}} \mid q, G^*) \cdot \mathcal{M}_{\theta1}(G^* \mid q, \mathcal{G})}_{\text{Graph Retrieval}},\nonumber
\end{split}
\end{equation}
where \( \mathcal{M}_{\theta1}(G^* \mid q, \mathcal{G}) \) models the distribution over subgraphs \( G^* \), given the query and full graph. \( \mathcal{M}_{\theta2} \) encodes local and global structure from the subgraph \( G^* \), producing representations suitable for downstream reasoning. \( \mathcal{M}_{\theta3} \) is an LLM-based module that conditions on the original query, the local quadruplets \( \mathcal{V}_q^{\text{local}} \), and the graph-encoded global representation \( \mathbf{h}_q^{\text{global}} \) to generate the final answer.

\paragraph{Finetuning Objective.}  
To capture temporal dynamics during fine-tuning, we integrate DG-LoRA into the LLM, enhancing standard LoRA with a timestamp-conditioned gating mechanism (see Eq.~(\ref{eq:DG-loRA}) in Section~\ref{dglora}). For each token representation $x_{\text{tok}}$ associated with timestamp $t$, the input to DG-LoRA is formed as $x = [x_{\text{tok}}; t]$, enabling the model to adjust adaptation strength based on temporal context. The low-rank update $\Delta W_{\text{DG}}$ is computed via Eq.~(\ref{eq:DG-loRA}), with the gating vector defined in Eq.~(\ref{eq:DG-loRA1}). This mechanism allows the model to adapt flexibly to varying temporal contexts.

The fine-tuning objective maximizes the log-likelihood of the ground-truth answer $\mathcal{A}_q$, conditioned on the query $q$, the DRGP-derived global summary $\mathbf{h}_q^{\text{global}}$, and the MI-filtered local subgraph $\mathcal{V}_q^{\text{local}}$:
\begin{align}
    \mathcal{L} = \mathbb{E}_{(q, \mathcal{V}_q^{\text{local}}, \mathbf{h}_q^{\text{global}}, \mathcal{A}_q) \sim \mathcal{D}} 
    \left[ \log P_{\theta + \Delta W_{\text{DG}}} \big( \mathcal{A}_q \big|
    q, \mathbf{h}_q^{\text{global}}, \mathcal{V}_q^{\text{local}} \big) \right],
    \nonumber
\end{align}
where $\theta$ denotes the frozen pretrained parameters and $\Delta W_{\text{DG}}(x,t)$ is the timestamp-conditioned low-rank adaptation. During training, only the DG-LoRA parameters $(W_A, W_B, W_C)$ are updated, enabling temporally grounded adaptation with minimal parameter overhead.

\section{Theoretical Analysis of DG-LoRA}

\subsection{Optimization Dynamics of DG-LoRA}\label{DG-LoRA}

We first analyze DG-LoRA from the perspective of gradient propagation, highlighting how the input-conditioned gate changes the optimization dynamics of low-rank adaptation parameters.

For standard LoRA in Eq.~\eqref{eq:LoRA}, the gradients of $W_A$ and $W_B$ with respect to the loss function $\mathcal{L}$ are
\begin{align}
\frac{\partial \mathcal{L}}{\partial W_A}
&= x^\top \frac{\partial \mathcal{L}}{\partial y} W_B^\top
\in \mathbb{R}^{d_1\times r},  
\label{eq:g-LoRA1} \\
\frac{\partial \mathcal{L}}{\partial W_B}
&= (x W_A)^\top \frac{\partial \mathcal{L}}{\partial y}
\in \mathbb{R}^{r\times d_2}.
\label{eq:g-LoRA2}
\end{align}

For DG-LoRA in Eq.~\eqref{eq:DG-loRA}, where
$y = xW + (xW_A \odot \text{gate}_c)W_B$ and
$\text{gate}_c = G(xW_C)$, the gradients are
\begin{align}
\frac{\partial \mathcal{L}}{\partial W_A}
&= x^{\top}\Big( 
\big( \tfrac{\partial \mathcal{L}}{\partial y} W_B^{\top} \big)
\odot \text{gate}_c 
\Big),
\label{eq:g-DG-LoRA1} \\
\frac{\partial \mathcal{L}}{\partial W_B}
&= \big(x W_A\odot \text{gate}_c\big)^{\top}
\Big(\tfrac{\partial \mathcal{L}}{\partial y}\Big),
\label{eq:g-DG-LoRA2} \\
\frac{\partial \mathcal{L}}{\partial W_C}
&= x^{\top}\Big( 
\big( \tfrac{\partial \mathcal{L}}{\partial y} W_B^{\top} \big)
\odot x W_A
\odot \mathbf{1}\{xW_C\ge0\} \notag\\
&\quad
\odot \big(1-\text{gate}_c\odot \text{gate}_c\big)
\Big).
\label{eq:g-DG-LoRA3}
\end{align}
Here, $\frac{\partial \mathcal{L}}{\partial y}\in \mathbb{R}^{S \times d_2}$ is the output gradient, 
$\frac{\partial \mathcal{L}}{\partial W_A}, \frac{\partial \mathcal{L}}{\partial W_C} \in \mathbb{R}^{d_1 \times r}$ are the gradients of the compression and gating matrices, and 
$\frac{\partial \mathcal{L}}{\partial W_B} \in \mathbb{R}^{r \times d_2}$ is the gradient of the reconstruction matrix.

Compared with standard LoRA, DG-LoRA introduces two optimization effects. 
First, $\text{gate}_c$ selectively scales the gradients of $W_A$ and $W_B$, so only activated low-rank components receive effective updates. 
This makes parameter optimization input-dependent and sparse. 
Second, $W_C$ receives gradients jointly determined by the reconstruction path, compressed features, and gate derivative, enabling it to learn which low-rank components should be activated under different inputs.

\subsection{Theoretical Justification for DG-LoRA}\label{proof}

We further interpret DG-LoRA as input-conditional low-rank adaptation. 
Given an input representation $x$ containing temporal information, DG-LoRA produces
\begin{align}
y &= xW + \Delta y, \\
\Delta y &= (xW_A \odot G(xW_C))W_B,
\end{align}
where $W$ is frozen, $W_A$ and $W_B$ are low-rank trainable matrices, and $W_C$ is the gating matrix. 
The gating function is
\begin{equation}
G(z)=\mathbb{I}(z\geq 0)\odot \tanh(z).
\end{equation}

\textbf{Input-conditional adaptation.}
Standard LoRA applies a fixed update:
\begin{equation}
\Delta y_{\mathrm{LoRA}}=xW_AW_B.
\end{equation}
In contrast, DG-LoRA can be rewritten as
\begin{equation}
\Delta y=xW_A D(x) W_B,
\quad
D(x)=\mathrm{Diag}(G(xW_C)),
\end{equation}
which induces an input-specific effective update:
\begin{equation}
\Delta W(x)=W_A D(x) W_B.
\end{equation}
Thus, DG-LoRA defines a family of input-conditional low-rank updates rather than a single static adapter. 
When $x$ includes temporal signals, the effective update becomes conditioned on timestamps or validity intervals.

\textbf{Adaptive effective rank.}
Let
\begin{equation}
\mathcal{A}(x)=\{i \mid G_i(xW_C)\neq 0\}
\end{equation}
be the active set of low-rank components. 
Since $D(x)$ is diagonal and sparse,
\begin{equation}
\mathrm{rank}(\Delta W(x))
=
\mathrm{rank}(W_A D(x) W_B)
\leq
|\mathcal{A}(x)|
\leq r.
\end{equation}
Therefore, DG-LoRA dynamically adjusts the effective rank according to input complexity, allocating fewer components to simple contexts and more capacity to temporally specific ones.

\textbf{Bounded sparse update.}
The indicator term $\mathbb{I}(z\geq 0)$ suppresses inactive components, while $\tanh(z)$ bounds active gates within $(-1,1)$. 
Thus, $\Delta W(x)$ is constrained by sparse and bounded diagonal modulation, reducing uncontrolled low-rank perturbations and improving robustness under uneven temporal data density.

Overall, DG-LoRA improves standard LoRA by replacing a globally shared low-rank update with an input-conditional effective update and by controlling its effective rank through sparse gating. 
For TKG reasoning, this enables different temporal contexts to activate different adaptation subspaces, thereby better modeling time-dependent entity semantics and relation drift.

\section{Experiments}
In our experiments, we aim to answer the following research questions: \textbf{RQ1:} Does \texttt{TEPCA} achieve state-of-the-art performance on temporal knowledge graph reasoning tasks? \textbf{RQ2:} How much does each component of \texttt{TEPCA} contribute to temporal reasoning performance? \textbf{RQ3:} How does the rank of \texttt{DG-LoRA} affect the balance between adaptation capacity and generalization? \textbf{RQ4:} How do fine-tuning iterations affect the convergence and stability of \texttt{TEPCA} compared with standard LoRA? \textbf{RQ5:} Does \texttt{DG-LoRA} exhibit general input-adaptive adaptation capacity across different LLM backbones and commonsense reasoning datasets? \textbf{RQ6:} Can the dynamic low-rank subspaces learned by \texttt{DG-LoRA} reveal interpretable input-dependent adaptation patterns? 

\subsection{Datasets}\label{dataset}
Table \ref{tab:label_data} reports dataset statistics, including the number of entities, relations, and temporal facts, as well as the granularity of each dataset. These statistics characterize scale and temporal resolution and support the evaluation of TEPCA across settings.

\begin{table}[h]
\centering
\caption{Statistical information of datasets.}
\setlength{\tabcolsep}{1.3mm}
\begin{tabular}{lcccccc}
\toprule
\textbf{Dataset} & \textbf{Entities} & \textbf{Relations} & \textbf{\#Train} & \textbf{\#Val} & \textbf{\#Test} & \textbf{Granularity} \\
\midrule
ICEWS14 & 6{,}869 & 230 & 74{,}845 & 8{,}514 & 7{,}371 & 24 hours \\
\rule{0pt}{9pt}
GDELT   & 5{,}850 & 238 & 79{,}319 & 9{,}957 & 9{,}715 & 15 mins  \\
\bottomrule
\end{tabular}
\label{tab:label_data}
\end{table}

Integrated Crisis Early Warning System (ICEWS) is a TKG for international crisis early warning and analysis, which collects and integrates data on political events, social dynamics, and international relations worldwide. ICEWS14 \cite{Learning2018} contains events that occurred from $1/1/2014$ to $12/31/2014$.
GDELT \cite{leetaru2013gdelt}, short for the Global Database of Events, Language, and Tone, is a large-scale TKG that records political and social events between geopolitical entities over time. Each record contains a timestamp, interacting entities, event types, and other contextual metadata. Due to its real-world coverage and long temporal span, GDELT serves as a challenging benchmark for temporal reasoning under high noise and data sparsity.

To further demonstrate DG-LoRA’s effectiveness, we evaluate it on eight NLP question-answering benchmarks curated by LLaMA-Adapter \cite{hu2023llm}: BoolQ \cite{clark2019boolq}, PIQA \cite{bisk2020piqa}, SIQA \cite{sap2019social}, HellaSwag \cite{zellers2019hellaswag}, WinoGrande \cite{sakaguchi2021winogrande}, ARC-e \cite{clark2018think}, ARC-c \cite{clark2018think}, and OBQA \cite{mihaylov2018can}. These benchmarks span reading comprehension and physical, social, and scientific commonsense, providing a diverse testbed for evaluating DG-LoRA’s dynamic low-rank subspace gating.

\begin{table*}[t]
    \centering
    \caption{Link prediction results on ICEWS14 and GDELT.
    Bold values indicate the best results; ``--'' denotes unavailable data.}
    \label{tab:my_label1}
    \setlength{\tabcolsep}{3mm}
    {
    \begin{tabular}{llccc c ccc}
    \toprule
    \multirow{2}{*}{Method Type}
    & \multirow{2}{*}{Methods}
    & \multicolumn{3}{c}{ICEWS14}
    & {}
    & \multicolumn{3}{c}{GDELT} \\
    \cline{3-5} \cline{7-9}
    & & Hit@1 & Hit@3 & Hit@10
    & & Hit@1 & Hit@3 & Hit@10 \\
    \midrule

    \multirow{7}{*}{Embed/Rule-based}
    & xERTE \cite{han2020explainable}
    & 0.330 & 0.454 & 0.570 & & 0.085 & 0.159 & 0.265 \\
    & RE-GCN \cite{li2021temporal}
    & 0.313 & 0.473 & 0.626 & & 0.084 & 0.171 & 0.299 \\
    & TLogic \cite{liu2022tlogic}
    & 0.332 & 0.476 & 0.602 & & 0.113 & 0.212 & 0.351 \\
    & TR-Rules \cite{li2023tr}
    & 0.339 & 0.485 & 0.611 & & -- & -- & -- \\
    & RE-H$^2$AN \cite{guo2023hierarchical}
    & 0.312 & 0.458 & 0.603 & & 0.114 & 0.199 & 0.331 \\
    & LGevo \cite{luo2025temporal}
    & 0.367 & 0.557 & 0.728 & & 0.129 & 0.219 & 0.361 \\
    & LogiQ \cite{chen2025enhancing}
    & 0.357 & 0.510 & 0.642 & & -- & -- & -- \\
    \midrule

    \multirow{12}{*}{LLMs-based}
    & PPT (BERT-cased) \cite{han2020explainable}
    & 0.273 & 0.408 & 0.563 & & -- & -- & -- \\
    & ChapTER (BERT-uncased) \cite{peng-etal-2024-deja}
    & 0.242 & 0.380 & 0.527 & & -- & -- & -- \\
    & HFL (BERT-uncased) \cite{xu2025historical}
    & 0.277 & 0.427 & 0.573 & & -- & -- & -- \\
    \cmidrule(lr){2-9}

    & ICL (InternLM2) \cite{lee2023temporal}
    & 0.301 & 0.432 & 0.560 & & 0.084 & 0.152 & 0.254 \\
    & ICL (Llama-2-7B) \cite{lee2023temporal}
    & 0.252 & 0.427 & 0.504 & & 0.060 & 0.164 & 0.246 \\
    & ICL (GPT-NeoX-20B) \cite{lee2023temporal}
    & 0.334 & 0.460 & 0.565 & & -- & -- & -- \\
    \cmidrule(lr){2-9}

    & CoH (Llama-2-7B) \cite{luo2024chain}
    & 0.349 & 0.470 & 0.591 & & -- & -- & -- \\
    & CoH (Vicuna-7B) \cite{luo2024chain}
    & 0.328 & 0.457 & 0.656 & & -- & -- & -- \\
    & CoH (Mixtral-8x7B) \cite{xia2024enhancing}
    & 0.331 & 0.496 & 0.649 & & -- & -- & -- \\
    \cmidrule(lr){2-9}
    & LLM-DA (GPT-3.5-turbo) \cite{wang2024large}
    & 0.356 & 0.515 & 0.662 & & 0.121 & 0.221 & 0.331 \\

    & GenTKG (GPT-NeoX-20B) \cite{liao2024gentkg}
    & 0.350 & 0.474 & 0.575 & & 0.102 & 0.167 & 0.273 \\
    & GenTKG (Llama-2-7B) \cite{liao2024gentkg}
    & 0.361 & 0.472 & 0.535 & & 0.139 & 0.225 & 0.304 \\
    \midrule

    & \textbf{TEPCA (Llama-2-7B)}
    & \textbf{0.371} & \textbf{0.559} & \textbf{0.768}
    & & \textbf{0.143} & \textbf{0.239} & \textbf{0.363} \\
    \bottomrule
    \end{tabular}
    }
\end{table*}

\subsection{Baselines}\label{baseline}

We compare TEPCA with representative baselines from three categories: 
traditional TKGR methods, PLM-based TKGR methods, and LLM-based TKGR methods. 
These baselines cover embedding-based temporal modeling, rule-based temporal reasoning, prompt-based temporal prediction, and fine-tuning-based LLM reasoning, providing a comprehensive comparison with both conventional and recent approaches.

\noindent\textbf{Traditional TKGR methods.}
We include representative embedding-based and rule-based TKGR methods, including RE-GCN~\cite{li2021temporal}, xERTE~\cite{han2020explainable}, TLogic~\cite{liu2022tlogic}, TR-Rules~\cite{li2023tr}, LGevo~\cite{luo2025temporal}, and LogiQ~\cite{chen2025enhancing}. 
These methods model temporal relational patterns through recurrent graph encoding, explainable temporal subgraph reasoning, temporal rule mining, evolutionary pattern modeling, or rule-guided reinforcement learning.

\noindent\textbf{PLM-based TKGR methods.}
We further compare with PLM-based temporal reasoning methods, including ChapTER~\cite{peng-etal-2024-deja}, PPT~\cite{xu2023pre}, and HFL~\cite{xu2025historical}. 
These methods formulate TKGR with pre-trained language models by incorporating timestamp semantics, temporal prompts, or historical fact modeling.

\noindent\textbf{LLM-based TKGR methods.}
We also evaluate against recent LLM-based TKGR methods, including InternLM2-ICL, Llama-2-7B-ICL, and GPT-NeoX-ICL~\cite{lee2023temporal}; Llama-2-7B-CoH and Vicuna-7B-CoH~\cite{luo2024chain}; Mixtral-8x7B-CoH~\cite{xia2024enhancing}; GPT-NeoX-20B-GenTKG and Llama-2-7B-GenTKG~\cite{liao2024gentkg}; and GPT3.5-turbo-LLM-DA~\cite{wang2024large}. 
These baselines represent in-context learning, chain-of-history reasoning, temporal logical rule-enhanced generation, and dynamic LLM adaptation for temporal prediction.




\subsection{Evaluation and Implementation}\label{ps}

\textbf{Evaluation Metrics.}
We follow the standard filtered temporal link prediction protocol and report Hit@N ($N=1,3,10$) as the main evaluation metrics. 
For each query, all candidate entities are ranked according to the model prediction, and other known valid facts appearing in the train, validation, or test split are filtered out before computing the final rank. 
Hit@N measures whether the ground-truth entity appears among the top-$N$ ranked candidates, thereby reflecting the model's ability to identify temporally valid missing entities under a given query timestamp.

\textbf{Entity Matching Strategy.}
Since LLM-based TKGR methods generate textual answers rather than entity IDs, we map generated outputs to candidate entities before computing link prediction metrics. 
For the main comparison, we adopt Longest Match, which deterministically selects the longest candidate entity string appearing in the generated output. 
This protocol ensures reproducible entity-level evaluation and avoids introducing an additional semantic encoder into the primary metric. 
Cosine Similarity Matching and Open Generation Matching are used only for auxiliary analysis: the former evaluates robustness to lexical variants and paraphrases but may over-credit semantically similar incorrect entities, whereas the latter provides a stricter surface-form matching protocol. 

\textbf{Backbone LLM.}
Unless otherwise specified, TEPCA uses Llama-2-7B as the backbone LLM for the main TKGR experiments. 
We choose Llama-2-7B for two reasons. 
First, it is an open-source backbone that has been widely adopted in prior LLM-based TKGR studies, enabling controlled comparison under comparable model capacity. 
Second, our goal is not to rely on increasingly larger LLMs, but to examine whether temporally aligned retrieval and timestamp-conditioned adaptation can compensate for the static-memory limitation of LLMs. 
Additional experiments on multiple LLM backbones are conducted to evaluate the generality of DG-LoRA beyond a single model architecture.

\textbf{Retrieval and Pruning Settings.}
For Query Expansion and DRGP pruning, the temporal decay coefficient is set to $\lambda=0.1$ by default. 
The maximum retrieval threshold is set to $k=20$ on both ICEWS14 and GDELT, which provides a balance between evidence coverage and prompt length. 
For MI-based pruning, the temperature parameter in semantic compatibility is set to $\tau=0.1$, and the temporal decay coefficient is set to $\lambda_2=0.1$ unless otherwise specified. 
In the background salience term $\rho(g)$, we set $\eta=0.5$ to balance entity-level structural salience and relation-level frequency, and use $\epsilon=10^{-6}$ for numerical stability. 
The normalized entity degree and normalized relation frequency are computed within the query-induced temporal candidate subgraph $\mathcal{G}_q^{-}$. 
All hyperparameters are selected according to validation performance.

\textbf{DG-LoRA Configuration.}
DG-LoRA is applied to the attention projection layers of the backbone LLM. 
The rank of DG-LoRA is selected from $\{4,8,16,32\}$ on the validation set. 
Accordingly, we set the DG-LoRA rank to 8 on ICEWS14 and 32 on GDELT, balancing temporal adaptability and model complexity. 
During fine-tuning, the original LLM parameters are frozen, and only the DG-LoRA parameters are updated. 
The maximum number of fine-tuning epochs is set to 4, and the final checkpoint is selected according to validation performance to avoid overfitting to temporally sparse intervals.

All experiments are implemented in PyTorch and conducted on four NVIDIA A100 GPUs with an Intel Xeon Gold 6326 CPU. 


\begin{table*}[t] 
    \centering
    \caption{Ablation study on ICEWS14. The best results are in bold. ``\textit{(-)}'' denotes replacing DG-LoRA with original LoRA. ``\textit{w/o}'' indicates ``without''.} 
    \setlength{\tabcolsep}{6mm}{
    \begin{tabular}{ c c c c c c c c}
    \toprule
    \rule{0pt}{5pt}
    \multirow{2}{*}{Methods}&\multicolumn{3}{c}{ICEWS14}&{}&\multicolumn{3}{c}{GDELT}\\
    \cline{2-4}\cline{6-8}
    {}&{Hit@1}&{Hit@3}&{Hit@10}&{}&{Hit@1}&{Hit@3}&{Hit@10}\\
    \hline
    {TEPCA \textit{(-)} LoRA}&{0.347}&{0.524}&{0.748}&{}&{0.132} & {0.220} & {0.342}\\
    {TEPCA \textit{w/o} QE}&{0.359}&{0.550}&{0.759}&{}&{0.135} & {0.228} & {0.347}\\
    {TEPCA \textit{w/o} DRGP}&{0.339}&{0.519}&{0.727}&{}&{0.124} & {0.213} & {0.332}\\
    {TEPCA \textit{w/o} MI}&{0.318}&{0.541}&{0.751}&{}&{0.118} & {0.219} & {0.344}\\
    \hline
    {TEPCA}&{\textbf{0.371}}&{\textbf{0.559}}&{\textbf{0.768}}&{}&{\textbf{0.143}}&{\textbf{0.239}}&{\textbf{0.363}}\\
    \bottomrule
   \end{tabular}}
\label{tab:my_label2}
\end{table*}

\subsection{RQ1: TKGR Performance Comparison}
The link prediction results for ICEWS14 and GDELT are presented in Table~\ref{tab:my_label1}. The experimental analyses are listed as follows:

(1) Experimental results indicate that the proposed TEPCA achieves state-of-the-art performance across all metrics on both ICEWS14 and GDELT, demonstrating its robustness in TKGR. This improvement is attributed to two key design choices: (i) the dynamic relevance-guided path pruning combined with mutual information–based pruning filters out irrelevant facts to retain only time-aligned, semantically rich paths from the TKGs; and (ii) the Dynamic-Gated LoRA (DG-LoRA) fine-tuning adapts the low-rank parameters of LLMs to query-specific temporal and semantic contexts. Together, these components enable TEPCA to better contextualize queries with temporally precise knowledge, especially in time-sensitive or multi-hop reasoning scenarios.

(2) TEPCA demonstrates clear advantages over existing paradigms. Embedding-based TKGRs, constrained by static representations, struggle to capture the evolving temporal dynamics of entities and relations. Rule-based approaches provide interpretability but suffer from limited flexibility and generalizability, especially in sparse or temporally ambiguous contexts. While LLM-based methods leverage powerful pretrained knowledge, they often rely on temporally misaligned evidence and uniform fine-tuning strategies that overlook the dynamic temporal roles of entities. In contrast, TEPCA integrates time-aligned retrieval with query-adaptive fine-tuning, enabling more precise and context-aware reasoning over TKGs.

(3) GDELT is a more challenging dataset for temporal link prediction than ICEWS14, with performance metrics for all methods being substantially lower. This disparity stems from its larger scale, higher sparsity, and complex event types and temporal dynamics, requiring robust and generalizable reasoning capabilities. TEPCA's state-of-the-art performance on GDELT, supported by comprehensive evaluation, demonstrates its robust and generalizable effectiveness in handling complex TKGs, thus addressing a gap in the literature.

(4) LLM-based methods generally outperform traditional TKGR methods. For example, while LGeVo attains a Hit@1 of 0.367 on ICEWS14, several LLM-based methods (e.g., LLM-DA and TEPCA) achieve comparable or higher scores. This indicates LLMs have an inherent advantage in leveraging their rich pre-trained knowledge and reasoning for TKG tasks. Nevertheless, LGeVo remains highly competitive, even surpassing some LLM-based methods (e.g., ICL and GenTKG on ICEWS14 Hit@1), suggesting well-engineered traditional methods are strong baselines and that substantial gains result from effectively integrating LLMs with TKG-specific mechanisms, as exemplified by TEPCA.

\begin{figure}[t]
    \centering
    \includegraphics[scale=0.46]{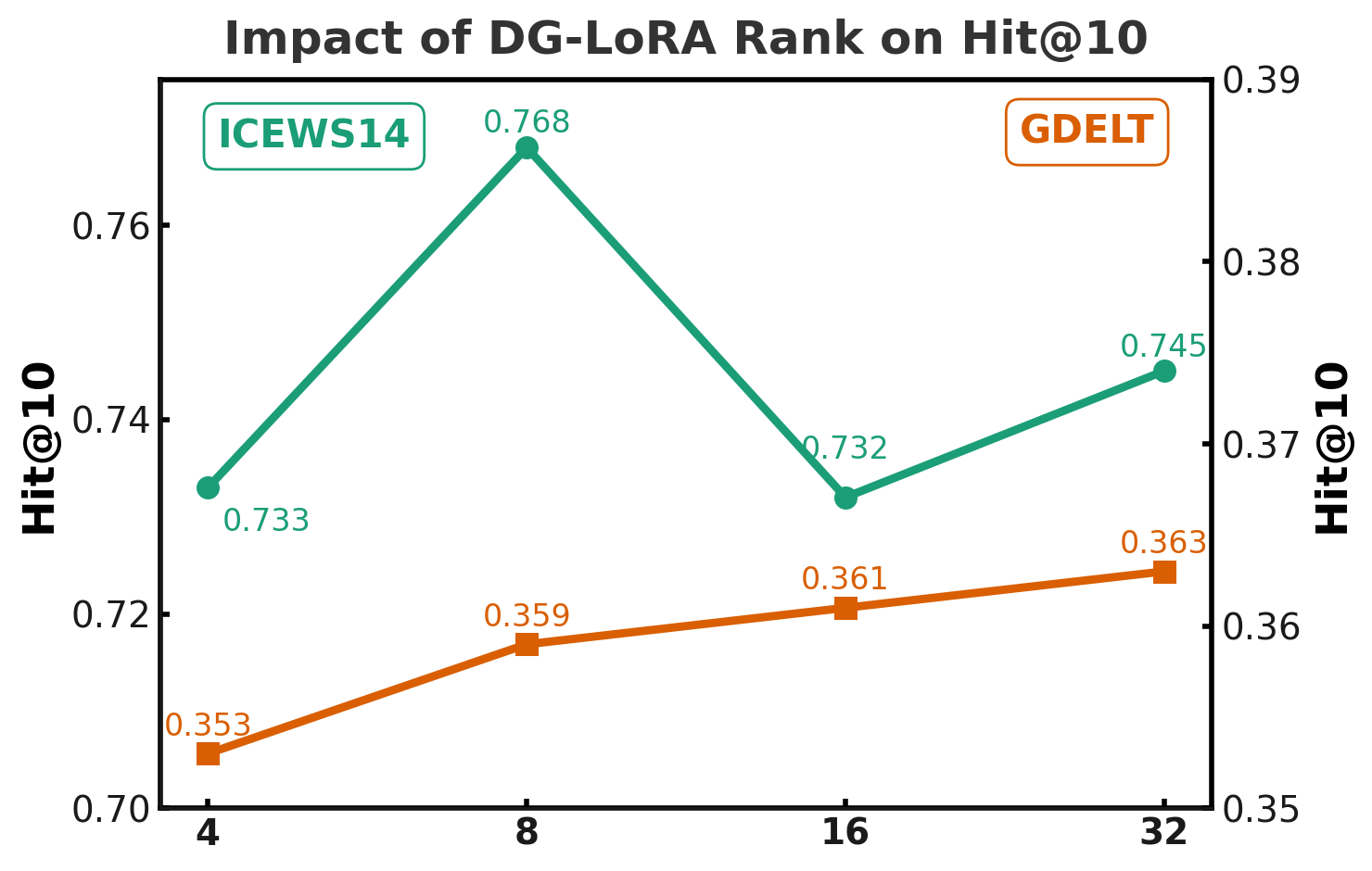}
    \caption{Impact of DG-LoRA rank on both datasets.
    }
    \label{fig:my_label_5}
\end{figure}

\begin{figure*}[t]
    \centering
    \includegraphics[scale=0.35]{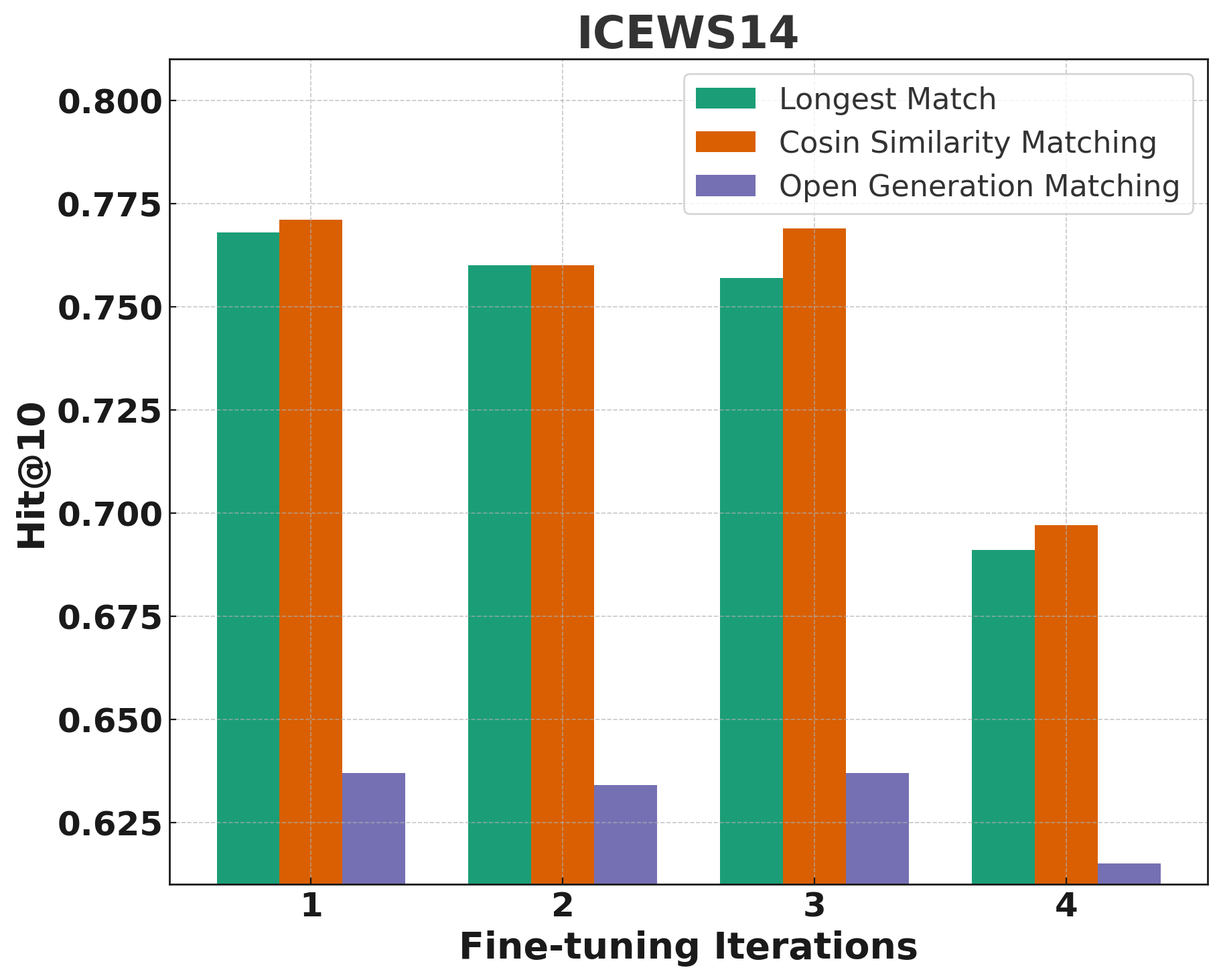}
    \quad\quad
    \includegraphics[scale=0.35]{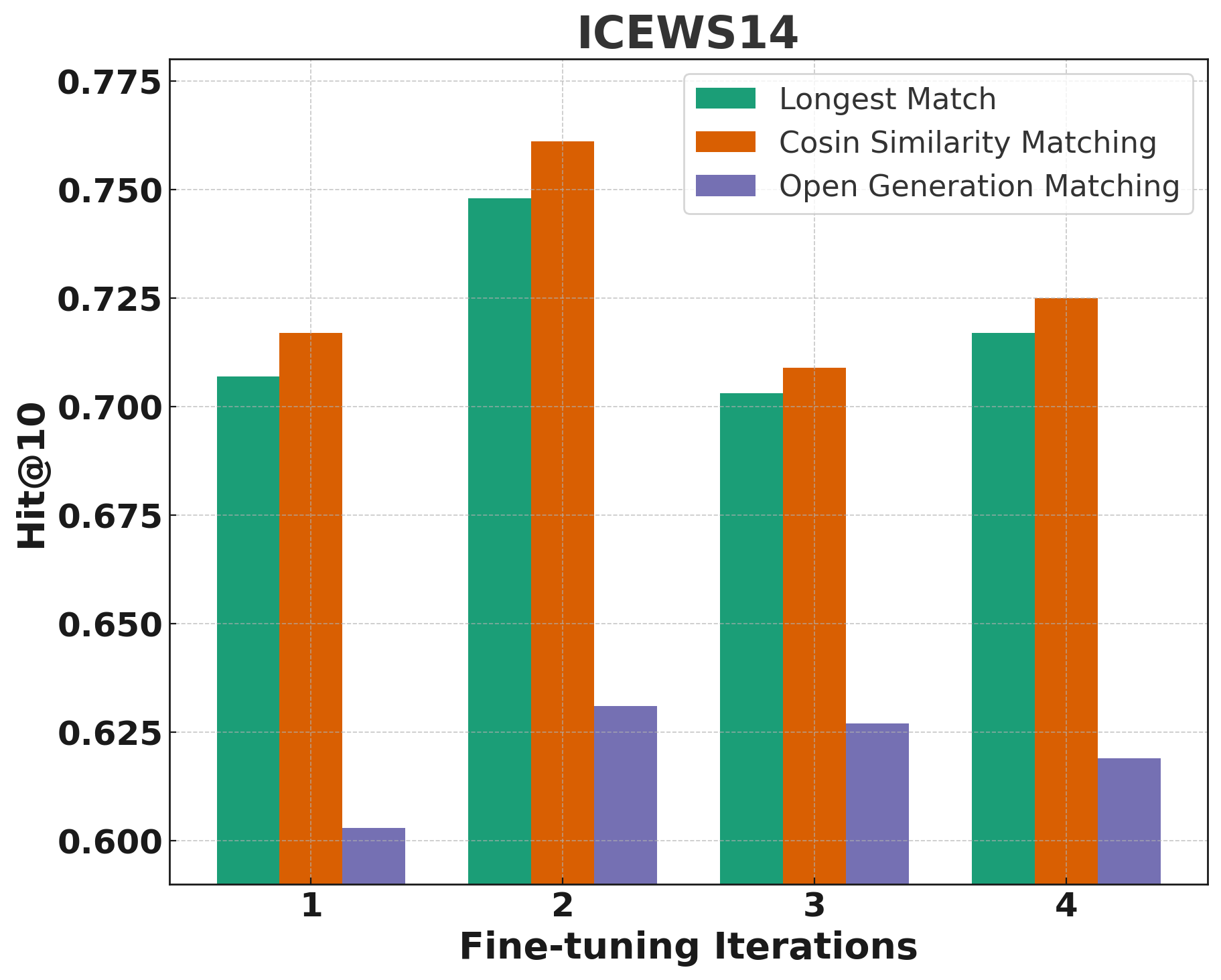}
    \caption{Comparison of varying numbers of fine-tuning on ICEWS14 dataset within Hit@10 under different matching strategies. The left subplot is ``\textit{TEPCA w/ DG-LoRA}'' model, and the right subplot is ``\textit{TEPCA w/ LoRA}'' model.
    }
    \label{fig:my_label_6v}
\end{figure*}

\begin{table*}[t]
    \centering
    \caption{Experimental results on eight commonsense reasoning datasets using various LoRA methods for Llama3.2-1B/3B, Llama2-7B, Qwen2.5-7B, and Llama3-8B. Notably, the rank value is 32 ($r=32$) and \textbf{best} indicates the highest average metric value among all methods.}
    \setlength{\tabcolsep}{3mm}{
    \begin{tabular}{llcccccccccc}
        \toprule
        LLMs & Methods & BoolQ & PIQA & SIQA & HellaSwag & WinoGrande & ARC-e & ARC-c & OBQA & \textbf{Avg.} \\
        \midrule
        ChatGPT & - & 73.1 & 85.4 & 68.5 & 78.5 & 66.1 & 89.8 & 79.9 & 74.8 & 77.0 \\
        \midrule
        \multirow{2}{*}{Llama3.2-1B} 
        & LoRA    & 63.9 & 75.7 & 72.0 & 64.7 & 67.4 & 71.0 & 53.6 & 67.2 & 66.9 \\
        & DG-LoRA (Ours) &63.7 	&75.9 	&72.9 	&74.1 	&68.4 	&71.1 	&54.9 	&68.2 &
 \cellcolor{bestcolor}\textbf{68.7} \\
        \midrule
        \multirow{2}{*}{Llama3.2-3B} 
        & LoRA   & 71.5 & 83.2 & 75.4 & 86.3 & 79.2 & 82.5 & 70.1 & 79.6 & 78.5 \\
        & DG-LoRA (Ours) &71.5 	&84.5 	&78.8 	&92.3 	&81.2 	&83.8 	&72.4 	&80.0 &	
 \cellcolor{bestcolor}\textbf{80.5} \\
        \midrule
        \multirow{2}{*}{Llama2-7B} 
        & LoRA    &69.8 	&79.9 	&79.5 	&83.6 	&82.6 	&79.8 	&64.7 	&81.0 	&77.6  \\
        & DG-LoRA (Ours) &71.9 	&80.4 	&80.7 	&90.6 	&84.4 	&79.6 	&68.9 	&80.8  	& 
 \cellcolor{bestcolor}\textbf{79.7} \\
        \midrule
        \multirow{2}{*}{Qwen2.5-7B} 
        & LoRA  & 68.4 & 88.8 & 80.3 & 94.6 & 84.8 & 95.3 & 88.6 & 89.6 & 86.3 \\
        & DG-LoRA (Ours)  &71.3 	&88.2 	&79.9 	&93.7 	&83.0 	&96.2 	&90.2 	&89.6 &
 \cellcolor{bestcolor}\textbf{86.5} \\
        \midrule
        \multirow{2}{*}{Llama3-8B} 
        & LoRA  & 72.2 & 87.4 & 79.8 & 94.6 & 84.9 & 88.6 & 77.6 & 86.0 & 83.9 \\
        & DG-LoRA (Ours) &75.7 	&89.1 	&81.7 	&95.8 	&85.6 	&91.5 	&81.1 	&86.4&
 \cellcolor{bestcolor}\textbf{85.8} \\
        \bottomrule
    \end{tabular}}  
    \label{tab:1}
\end{table*}

\begin{figure*}[t]
\centering
  \includegraphics[width=2\columnwidth]{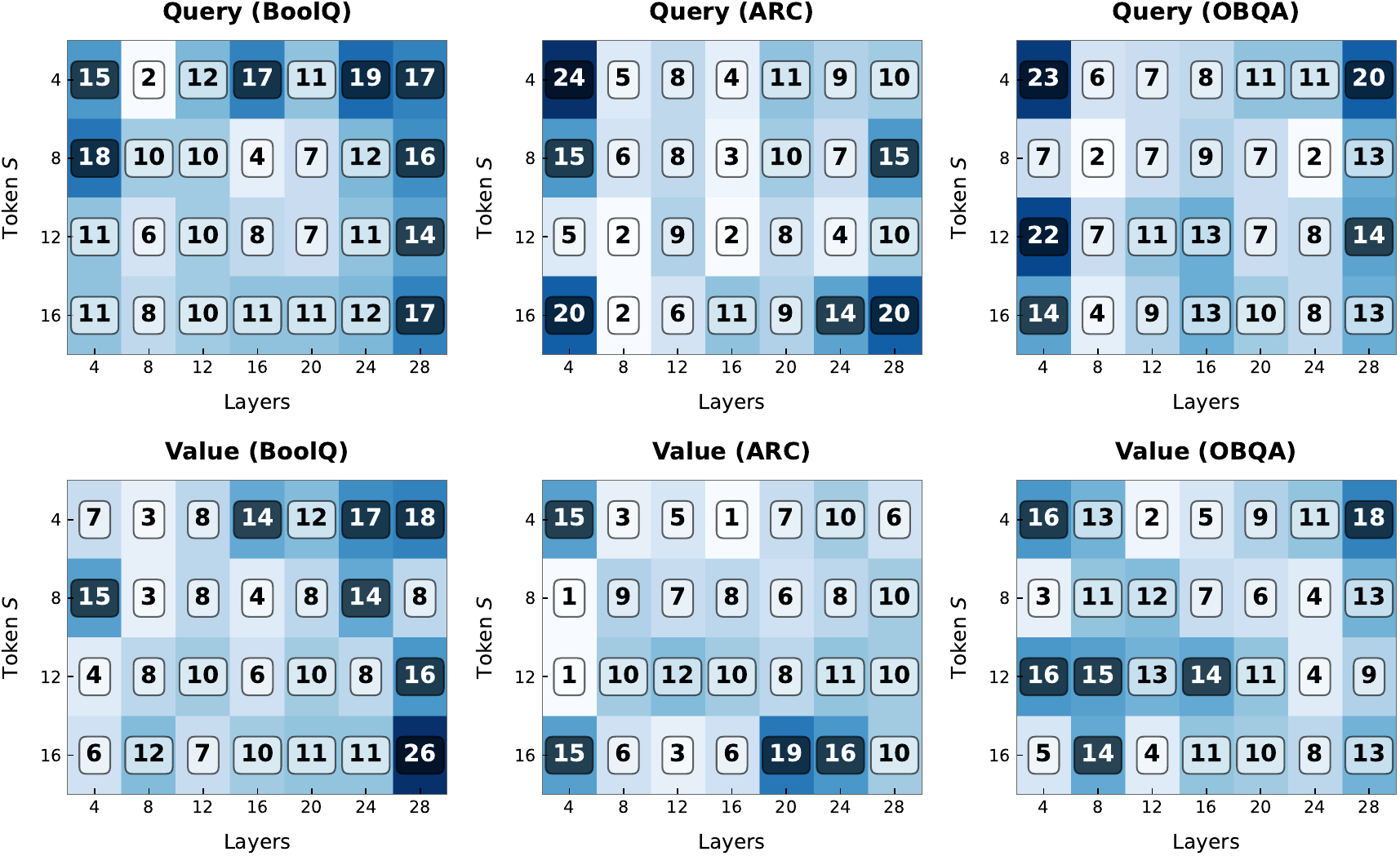}
  \caption{The low-rank subspace heatmap (the number of $\text{gate}_c\neq0$) of the Query ($q_{proj}$) and Value ($v_{proj}$) mappings for ARC, BoolQ, and OBQA question texts, based on the LLaMA-3.2-3B models ($r=32$).}
  \label{fig:4}
\end{figure*}

\subsection{RQ2: Component Effectiveness}
The ablation study on ICEWS14 and GDELT aims to evaluate the impact of different modules on temporal reasoning performance. 
We compare four variants of TEPCA by individually removing key components: the Dynamic Relevance-Guided Path (DRGP) pruning, the Mutual Information (MI)-based pruning, the Query Expansion (QE) module, and the Dynamic-Gated LoRA (DG-LoRA). 
The experimental results are shown in Table~\ref{tab:my_label2}.

\textit{Impact of DRGP Pruning.} 
Removing the DRGP module (\textit{TEPCA w/o DRGP}) leads to a noticeable decline in Hit@10 on both datasets, confirming its importance in capturing temporally consistent multi-hop reasoning paths that support LLM-based inference. 
While the drop in Hit@1 is modest, the larger decrease in Hit@3 and Hit@10 suggests that DRGP enhances the ranking quality of candidate answers by modeling global temporal structures rather than merely refining local evidence.

\textit{Impact of MI-based Pruning.} 
The absence of MI-based pruning (\textit{TEPCA w/o MI}) leads to the lowest Hit@1 across all variants, despite outperforming the DRGP ablation on Hit@3 and Hit@10. 
This observation indicates that the MI module plays a pivotal role in selecting highly informative and query-specific evidence, which is particularly crucial for accurate top-1 predictions. Without this mechanism, the retained local subgraph is likely to include semantically weaker or noisy facts, thereby undermining precise answer localization even when the global context remains intact.

\textit{Impact of Query Expansion.} 
Removing the query expansion module (\textit{TEPCA w/o QE}) results in consistent performance drops across all metrics, especially on GDELT. This underscores the importance of temporal context enrichment in disambiguating entity roles and enhancing both global path construction and local evidence selection.

\textit{Impact of DG-LoRA.} 
Replacing DG-LoRA with the standard LoRA configuration (\textit{TEPCA (-) LoRA}) results in consistent performance deterioration, validating the necessity of temporal-adaptive fine-tuning.
This confirms that entity semantics evolve over time, and fixed-rank adaptations are inadequate for capturing such temporal variability. 
DG-LoRA dynamically adjusts the low-rank update capacity based on timestamp-aware signals, enabling the model to better specialize in temporally diverse entity roles and relation patterns.

These ablation results collectively demonstrate the complementary contributions of the proposed components. 
DRGP provides global temporal structure, MI-based pruning ensures fine-grained relevance, query expansion enriches the contextual scope, and DG-LoRA enhances temporal adaptability. 

\subsection{RQ3: Rank Sensitivity}

To examine the impact of the DG-LoRA rank on reasoning performance, we vary the rank parameter \( r \in \{4, 8, 16, 32\} \) and evaluate Hit@10 on ICEWS14 and GDELT. As shown in Figure~\ref{fig:my_label_5}, increasing the rank from 4 to 8 leads to substantial gains on both datasets. Specifically, Hit@10 on ICEWS14 rises from 0.733 to 0.768, while GDELT improves from 0.353 to 0.359. This demonstrates that increasing the rank enhances the adaptation capacity, allowing it to better capture temporal variations in entity behavior.

However, further increasing the rank to 16 results in a notable performance drop on ICEWS14, while GDELT exhibits only marginal improvement. This suggests that although higher ranks provide greater expressiveness, they may also introduce overfitting, particularly in temporally dense datasets like ICEWS14.
Interestingly, setting the rank to 32 partially restores performance on ICEWS14 and yields continued but minor improvement on GDELT. These results indicate that excessively high ranks offer diminishing returns and may even degrade performance due to overparameterization.

These observations suggest that moderate DG-LoRA ranks strike the best balance between expressiveness and generalization.

\subsection{RQ4: Fine-tuning Dynamics}
\label{sec:exp_iterations}

Figure~\ref{fig:my_label_6v} illustrates the temporal reasoning performance (Hit@10) on the ICEWS14 dataset across varying fine-tuning iterations. We compare our proposed ``\textit{TEPCA w/ DG-LoRA}'' (Figure~\ref{fig:my_label_6v}, left) against the standard ``\textit{TEPCA w/ LoRA}'' (Figure~\ref{fig:my_label_6v}, right) using three evaluation protocols: \textit{Cosine Similarity Matching}, \textit{Longest Match}, and \textit{Open Generation Matching}.

\textit{Effectiveness of Matching Strategies.} 
\textit{Cosine Similarity Matching} consistently outperforms \textit{Longest Match} and \textit{Open Generation Matching} across all experimental settings. This superiority suggests that embedding-based evaluation better captures the semantic proximity between predicted entities and ground truths, avoiding the limitations of rigid exact string matching. Conversely, the consistently lower scores for \textit{Open Generation Matching} highlight the strictness of standard TKGR metrics, where minor lexical variations often result in false negatives.

\textit{Convergence and Stability of DG-LoRA.} 
Analyzing the training dynamics of \textit{DG-LoRA} (Figure~\ref{fig:my_label_6v}, left) reveals rapid convergence. The model achieves near-optimal performance at the first iteration and maintains stability through the third. This indicates that the time-aware gating mechanism facilitates swift alignment with temporal patterns without requiring extensive weight updates. However, a sharp performance degradation occurs at the 4-th iteration. This suggests that prolonged fine-tuning induces overfitting, where dynamic low-rank adapters memorize training noise at the expense of generalizability to unseen timestamps.

\textit{Superiority over Standard LoRA.} 
A comparison of the two subplots underscores the structural advantage of our approach. The baseline \textit{LoRA} (Figure~\ref{fig:my_label_6v}, right) exhibits a slower, more volatile convergence trajectory, peaking only at the 2-nd iteration with a lower maximum score. Unlike the significant oscillations observed in standard LoRA, \textit{DG-LoRA} not only converges faster but also sustains a higher performance plateau. This confirms that the dynamic gating mechanism effectively regulates adaptation, preventing the ``temporal specificity collapse'' common in static adapters and ensuring more reliable temporal reasoning.

\subsection{RQ5: Generality across LLM Backbones}\label{Evaluating DG-LoRA}

To further illustrate the effectiveness of DG-LoRA, we conduct further rich experiments on the 8 question-answering task of NLP field. As shown in Table~\ref{tab:1}, DG-LoRA consistently outperforms standard LoRA across all evaluated LLMs and most commonsense reasoning datasets. Notable gains are observed on LLaMA2-7B, where DG-LoRA achieves an average accuracy of 79.7\%, surpassing the 77.6\% obtained by standard LoRA. Significant improvements are also observed on specific tasks, such as BoolQ (+2.1), HellaSwag (+7.0), and ARC-c (+4.2). In addition, DG-LoRA yields consistent improvements over standard LoRA on LLaMA3.2-1B (+1.8), LLaMA3.2-3B (+2.0), and LLaMA3-8B (+1.9). These improvements can be attributed to DG-LoRA’s dynamic adaptation mechanism, which employs a learnable gating function $\text{gate}_c$ conditioned on temporal and contextual signals. 

Unlike standard LoRA that uses static low-rank updates shared across all inputs, DG-LoRA introduces a token-wise sparse gating controller (Eq.~\ref{eq:DG-loRA}--\ref{eq:DG-loRA1}) to modulate the contribution of low-rank components dynamically. This allows the model to specialize its reasoning for different contexts and semantic patterns, thereby enhancing generalization and representation flexibility. By adaptively discovering input-dependent low-rank subspaces, DG-LoRA mitigates the over-sharing limitation of traditional LoRA and demonstrates superior performance across diverse and challenging commonsense reasoning tasks.

\subsection{RQ6: Interpretability of Dynamic Subspaces} \label{Visualization}
The visualization of Dynamic-Gated Low Rank (DG-LoRA) effective low-rank subspaces reveals dynamic adjustments across both layers and tokens, providing empirical validation for the adaptive mapping mechanisms. As illustrated in Figure \ref{fig:4}, the dimensions of token-specific subspaces demonstrate substantial variation across different layers, confirming the filter controller's capacity to identify optimal low-rank subspaces during the compression process. Furthermore, the low-rank subspace dimension distribution across the three datasets (BoolQ, ARC, and OBQA) exhibits significant diversity at both token and layer levels. This observed diversity substantiates our core argument that DG-LoRA can adaptively learn the dynamic low rank subspace based on the different input features in diverse dataset fields.

\section{Conclusion}\label{conclusion}

In this paper, we propose a novel \textbf{T}emporal \textbf{E}vidence-\textbf{P}arameter \textbf{C}o-\textbf{A}lignment (\textbf{TEPCA}) framework that integrates Temporal Knowledge Graphs (TKGs) with Large Language Models (LLMs) to enhance both temporal reasoning and interpretability. We introduce two complementary pruning strategies: a time-constrained \textit{dynamic relevance-guided path pruning} that captures global semantic and temporal structure, and \textit{mutual information-based pruning} that extracts query-specific, temporally aligned subgraphs. Additionally, we propose Dynamic-Gated LoRA (DG-LoRA), a time-aware parameter-efficient tuning mechanism that adaptively modulates low-rank adaptation based on temporal context, enabling LLMs to better capture entity evolution and time-sensitive semantics. 

One limitation is that the timestamp-specific gating within DG-LoRA may overfit during periods with sparse data, leading to the memorization of noise from low-sample intervals. A promising avenue for future work could involve exploring temporal data augmentation, multi-scale regularization, and mixture-of-experts architectures to improve robustness across both sparse and dense time periods. 

\bibliographystyle{IEEEtran}
\bibliography{Ref}

\end{document}